\documentclass{bmvc2k}

\usepackage{graphicx}
\usepackage{amsmath}
\usepackage{amssymb}
\usepackage{subfigure}
\usepackage{floatrow}

\title{An Octree-Based Approach towards Efficient Variational Range Data Fusion}

\addauthor{Wadim Kehl}{kehl@in.tum.de}{ 1}
\addauthor{Tobias Holl}{holl@in.tum.de}{ 1}
\addauthor{Federico Tombari}{tombari@in.tum.de}{ 12 }
\addauthor{Slobodan Ilic}{slobodan.ilic@siemens.com}{ 13}
\addauthor{Nassir Navab}{navab@cs.tum.edu}{ 1}

\addinstitution{
 Computer-Aided Medical Procedures, \\
 TU Munich, Germany
}

\addinstitution{
 Computer Vision Lab (DISI), \\
 University of Bologna, Italy
}

\addinstitution{
 Siemens AG \\
 Research \& Technology Center \\
 Munich, Germany
}

\runninghead{Kehl et al.}{Octree-based Variational Range Data Fusion}

\def\eg{\emph{e.g}\bmvaOneDot}

\DeclareMathOperator*{\sgn}{sgn}
\DeclareMathOperator*{\divergence}{div}
\DeclareMathOperator*{\leafs}{leafs}

\newfloatcommand{capbtabbox}{table}[][\FBwidth]

\begin{document}

\maketitle

\begin{abstract}
Volume-based reconstruction is usually expensive both in terms of memory consumption and runtime. Especially for sparse geometric structures, volumetric representations produce a huge computational overhead. We present an efficient way to fuse range data via a variational Octree-based minimization approach by taking the actual range data geometry into account. We transform the data into Octree-based truncated signed distance fields and show how the optimization can be conducted on the newly created structures. The main challenge is to uphold speed and a low memory footprint without sacrificing the solutions' accuracy during optimization. We explain how to dynamically adjust the optimizer's geometric structure via joining/splitting of Octree nodes and how to define the operators. We evaluate on various datasets and outline the suitability in terms of performance and geometric accuracy. 
\end{abstract}

\section{Introduction}

\begin{figure}
\centering
\includegraphics[width=12cm]{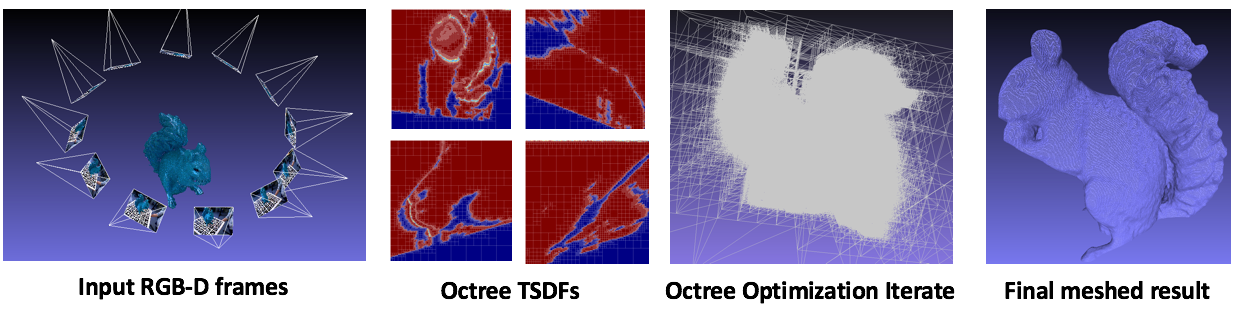}
\caption{Our work deals with robust variational fusion of range scans. Given a sequence of input frames, we optimize over Octree-structures representing transformed input TSDFs into a common, constantly evolving Octree to finally retrieve a geometrically accurate and smoothed meshed reconstruction.}
\label{fig:teaser}
\end{figure}

3D object reconstruction has been the focus of computer vision for decades and is important for many different fields including manufacturing verification, reverse engineering, rapid prototyping or augmented reality. Recently, the introduction of time-of-flight sensing devices as well as advances in stereo vision and especially the advent of low-cost RGB-D cameras further widened the focus and eased the reconstruction of objects and environments. 

Depending on the application the provided range maps are usually either fused into sparse clouds or into volumetric representations. While former are usually faster to store and process, the latter offer the advantage of straight-forwardly defining mathematical operators on the volumes due to the their dense nature. In this paper, we conduct 3D object reconstruction by fusing truncated signed distance fields (TSDF) via a variational optimization approach which would usually require us to store the data in dense volumetric grids. Such a representation entails a high memory consumption and runtime complexity since each discrete voxel has to be stored and also considered during operations. Furthermore, iterative optimization schemes create constantly changing intermediate solutions and thus would require for a dynamic spatial structure that follows the evolving implicit surface by structural updates. 

Hierarchical space partitioning schemes, like kD-trees \cite{Bentley1975} or Octrees \cite{Meagher1982}, can tremendously increase query performance for static geometries but need to be properly updated for dynamic changes occurring inside the volume. Obviously, employing partitioning schemes during an optimization must be carefully designed to avoid accumulation of quantization errors which lead to inaccurate solutions. In this work, we build our optimization around Octrees which recursively divide up the space into eight equally-sized cubes according to split and join rules. Octrees are established in many fields and are often used to alleviate computational burdens. Recent work (\cite{Steinbrucker2013,Chen2013,Zeng2013}) uses Octrees for range data integration to map the environment, but employs simple update rules to encompass newly seen data without any optimization whatsoever. For simulation problems these partitioning structures are usually of static auxiliary nature (\eg accelerating point/surface look-ups, \cite{Calakli2011,Losasso2004}) and get discarded or recomputed after each iteration. 

Our novel contribution is to use an Octree as the main optimization structure instead and we present a viable way to robustly fuse TSDFs in a variational approach, similar to \cite{Zach2007,Kehl2014}, while dynamically adapting the Octree's iterative structure to be faster and memory-efficient. We address the actual creation of the Octrees given initial range maps, the proper definition and calculation of mathematical quantities as well as correct numerical updates that include the iterative reorganization of the solution's hierarchical partitioning after each step. 

\section{Related work}

In many fields, level-set methods are often employed to solve given problems in e.g. fluid dynamics \cite{Bargteil2006,Losasso2004,Losasso2006}, computer graphics \cite{Baerentzen2002,Houston2006} or 3D reconstruction \cite{Curless1996,Kolev2012,Newcombe2011} where the physical properties of the model act upon the level-set function via PDEs. We refer the reader to a survey on 3D distance fields as a special variant of level-set functions \cite{Jones2006}. In contrast to explicit representations which can entail topological difficulties as well as rendering mathematical operations harder to implement, level-sets can implicitly represent arbitrary shapes and are therefore often preferred. Nonetheless, optimizing in volumetric data always is costly and related work tackled it in the following way: 

Similarly to our approach, \cite{Zach2007} also conducts variational data fusion and uses a runlength encoding that allows for fast decompression of the input data but does not address the problem of the structurally-changing minimizer during optimization. In a follow-up work \cite{Zach2008}, the authors propose a coarser quantization of the TSDF values and introduce a point-wise histogram based problem. In \cite{Schroers2012} the authors suggest to modify the data term such that one tries to be similar to the point-wise median. Again, none address the problem of the structurally-changing iterate.
The authors of \cite{Popinet2003} claim to have the first single-pass hierarchical-based approach for incompressible Euler equations based on multi-grids, although earlier works (e.g. \cite{Strain1999}) already focused on tracking moving interfaces with tree-based structures. In \cite{Losasso2004,Losasso2006}, the authors deal with spatially adaptive techniques for incompressible flow and state their surprise about the high accuracy of Octree-based mesh refinement even for small-scale structures.
The works \cite{Steinbrucker2013,Chen2013,Zeng2013} use dynamic Octree-based representations to store scene geometry but do no conduct any elaborate schemes for the integration since their main interest is mapping and efficient storage/updates for large-scale problems. \cite{Houston2006} introduces a data structure that includes a hierarchical partitioning where each cube uses a runlength encoding. Although very efficient in storage and lookup, online restructuring is rather slow and therefore less suited for iterative structural changes. In \cite{Niessner2013} and later \cite{Klingensmith2015}, the authors present an alternative to hierarchical partitioning by hashing the TSDF geometry to enable near-constant time lookups. All of the above mentioned works are focusing only on efficient storage and processing while paying little to no attention to the actual reconstruction quality. We instead are interested in reconstructing objects and thus on how to fuse the range data such that we perform efficiently in terms of memory and runtime while keeping a high reconstruction accuracy.

\section{Methodology}
Firstly, we show the computation of TSDFs with provided range data and the transformation into their Octree-representations. Secondly, we introduce the new problem formulation and give a way to efficiently solve it via node-wise split/join rules and a fast traversal technique.

Note that in contrast to \cite{Kehl2014} which fuses into RGB-D volumes, we solely focus here on the geometric minimizer because the space partitioning is not applicable to color volumes. Instead, we compute the coloring after meshing of the 0-isosurface similarly to \cite{Whelan2013}.


\begin{figure}
\begin{floatrow}

\ffigbox{ 
\includegraphics[width=3cm, height=3cm]{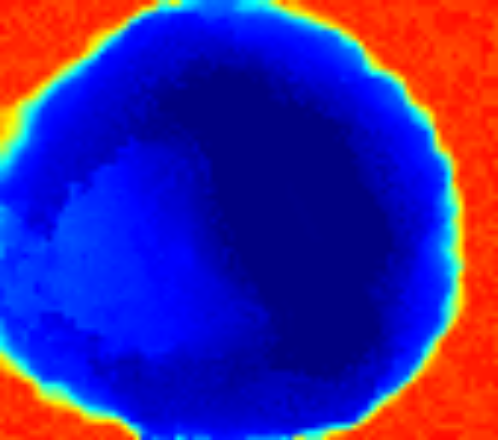}
\includegraphics[width=3cm, height=3cm]{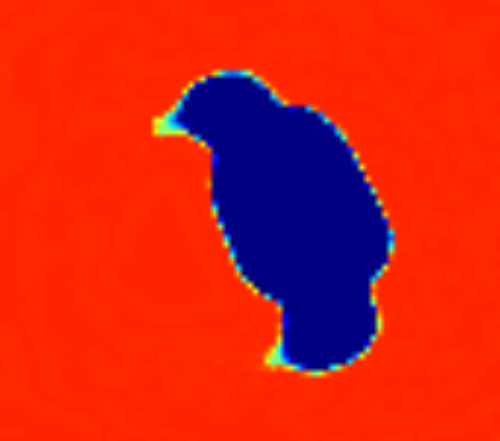}
}
{	\label{fig:delta}	\caption{Slicing through a dense TSDF with a large $\delta=20$ cm (left) and a very tight $\delta=2$ mm (right). The narrow band at the real object surface is clearly visible in the right image. We want to numerically focus on this interface while neglecting the uniform areas.}}

\ffigbox{
\includegraphics[width=3cm, height=3cm]{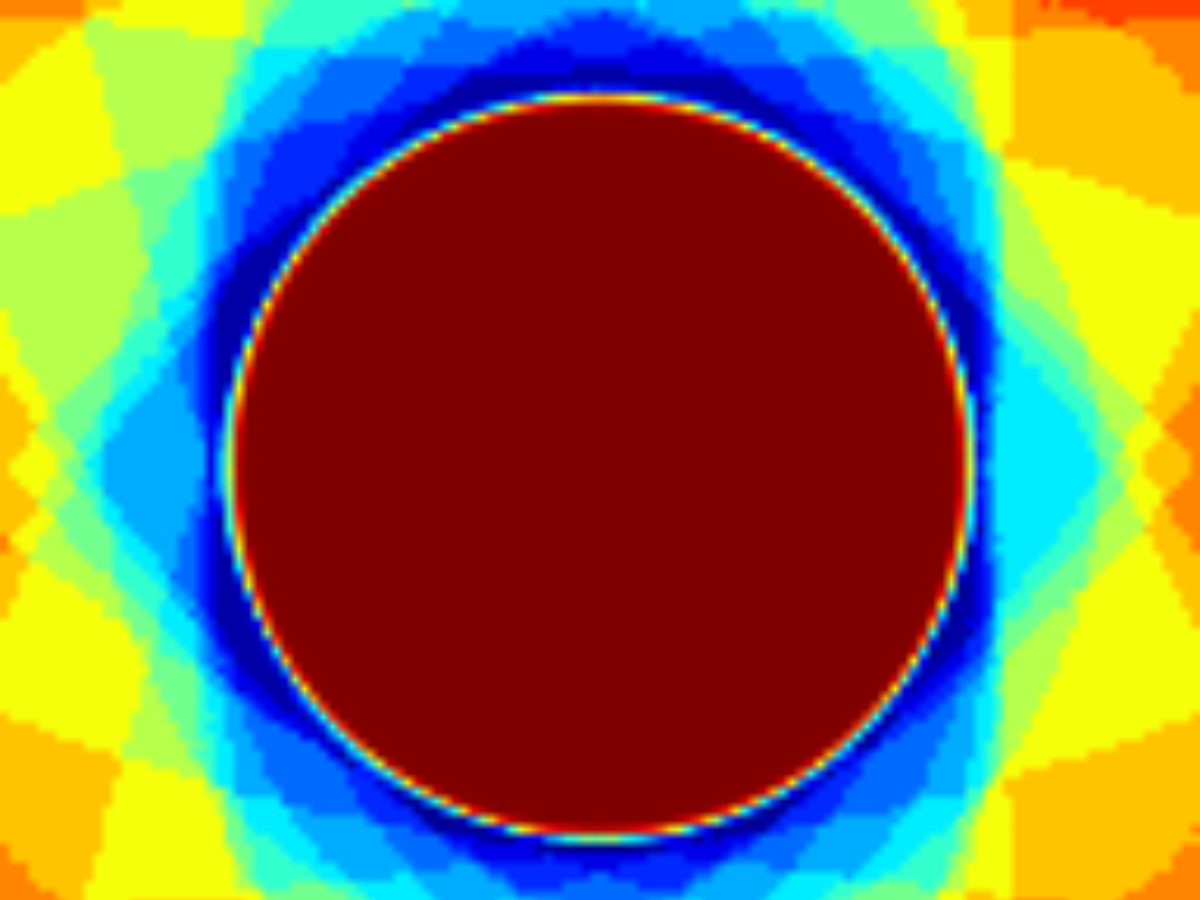} \includegraphics[width=3cm, height=3cm]{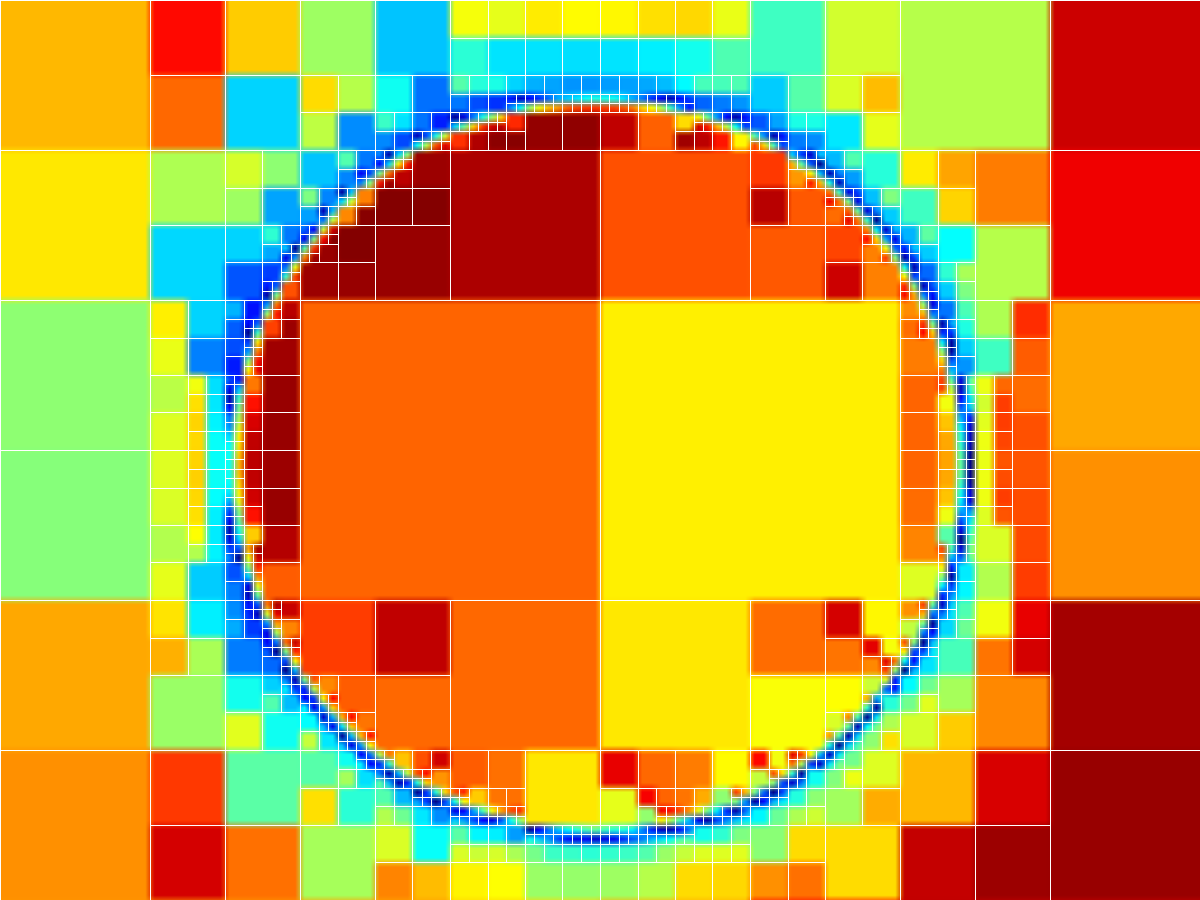}
}
{	\label{fig:partition}	\caption{Left: Slicing through a dense TSDF (left) and its Octree-version (right). Blue areas close to the surface possess a finer Octree-resolution since these represent the narrow band during optimization and should therefore be similar to the real TSDF values.}}	

\end{floatrow}
\end{figure}

\subsection{TSDF-Octree generation}
The generation of the TSDFs $f_i: \Omega_3 \subset \mathbb{R}^3 \rightarrow \mathbb{R}$ follows related work \cite{Zach2007,Schroers2012,Steinbrucker2013,Kehl2014}: given  the $i$-th frame of range data $D_i: \Omega_2 \rightarrow \mathbb{R}^+$ together with the corresponding projection $\pi_i: \mathbb{R}^3 \rightarrow \Omega_2$, the idea is to compute the signed distance $\phi_i$ between the surface and each point $\mathbf{x} \in \Omega_3$ of the reconstruction volume along the line of sight. Furthermore, scaling with $\delta$ and truncation to $[-1,+1]$ is performed to retrieve the final TSDF $f_i$

\begin{equation}
f_i(\mathbf{x}) = 
\begin{cases} 
\sgn(\phi_i(\mathbf{x})) &\mbox{if } |\phi_i(\mathbf{x})|>\delta \\ 
\phi_i(\mathbf{x}) / \delta & \mbox{else}  
\end{cases} \hspace{0.25cm}, \hspace{0.25cm} \text{with   } \phi_i(\mathbf{x}) = D_i(\pi_i(\mathbf{x})) - ||\mathbf{x}-C_i||
\end{equation} 
where $C_i$ is the projection center for the $i$-th frame in global coordinates. Additionally, we compute for each TSDF a weight volume $w_i:\Omega_3 \rightarrow \{0,1\}$ that signifies whether a voxel $\mathbf{x}$ has been observed from view $i$ or not via a check $\phi_i(\mathbf{x}) < -\eta$. Every unseen voxel in front of this hard $\eta$-threshold is assumed to be solid geometry for a given $f_i$. 

Above scaling factor $\delta$ serves as an uncertainty band and should be well-chosen both to compensate for measurement noise and to clearly divide between outer and interior space (see Figure \ref{fig:delta}). Strictly speaking, areas which are far apart from the interface consume memory and runtime during optimization without having a drastic influence on the optimizer's object surface. Our goal is to ensure that these spaces remain computationally inexpensive at all times during the optimization without impairing the final solution. To this end, we transform our problem to work with space-partitioned entities which can adapt to the changing narrow band of our iterated solutions.

\paragraph{Octree construction} We construct TSDF-Octrees $f_i^*$ from $f_i$ in a top-to-bottom manner. Starting from root node $n$, we define the spread $s$ of values subsumed by node $n$ in $f$ as
\begin{equation} 
s_f(n) = \bigg | \max_{\mathbf{x} \in \Omega_3(n)} f(\mathbf{x}) -\min_{\mathbf{x} \in \Omega_3(n)} f(\mathbf{x}) \bigg |
\end{equation}    
with $\Omega_3(n)$ being the subvolume that node $n$ represents. Initially, $\Omega_3(n) = \Omega_3$ and the spread will be maximal. From here we recursively apply a splitting rule: if the spread $s_f(n)$ at node $n$ is higher than a threshold $\tau = 0.1$, we subdivide $n$ into eight children and proceed further down. This is recursively repeated as long as the condition is fulfilled or until we reach a maximum Octree depth $D_{\max}$, corresponding to a pre-defined minimum metric voxel size. After the partitioning we propagate the means upwards from the leafs to all inner nodes to speed up computations during later optimization. See Figure \ref{fig:partition} for a visual comparison.

\subsection{Octree-based variational optimization}
Similar to \cite{Zach2007,Schroers2012,Kehl2014} we fuse all range maps into one volume by finding the minimizer of
\begin{equation} 
\mathcal{E}(u) := \int_{\Omega_3} D(\mathbf{f},\mathbf{w},u) + \lambda S(\nabla u)  \hspace{0.1cm} d \mathbf{x}
\end{equation}  
where we weight data fidelity against a regularization with a smoothness parameter $\lambda$. \cite{Zach2007} employs an outlier-robust $L^1$ data term $D(\mathbf{f},\mathbf{w},u) := \sum_i w_i \cdot |u - f_i|$ together with a Total Variation (TV) regularizer $S(\nabla u) := | \nabla u|$ which has the advantage of penalizing the perimeter of the level sets in $u$ and in combination, TV-$L^1$ induces a pure geometric regularization, as found in \cite{Chan2006}. Unfortunately, the non-differentiability requires elaborate solving schemes and we thus follow \cite{Schroers2012,Kehl2014} by tightly approximating both terms with differentiable quantities
\begin{equation}
D(\mathbf{f},\mathbf{w},u)  :=  \sum_i w_i \cdot \Gamma((u-f_i)^2)  \hspace{0.5cm},  
\hspace{0.5cm} S(\nabla u)  := \Gamma( | \nabla u|^2) 
\end{equation}
with $\Gamma(x^2) := \sqrt{x^2+\epsilon^2}$ together with a small $\epsilon > 0$ being the $epsL^1$ approximation \cite{Lee2006}.

An important aspect is the data term normalization since in its current form the functional puts more emphasis on the data term if the number of range images increases. Instead of dividing by the number of images we divide point-wise by the accumulated weight to achieve a spatially consistent smoothing, regardless of how often a voxel has been seen \cite{Schroers2012}.
Altogether, this strictly-convex functional can now be solved by gradient descent and in our Octree-based variant, we furthermore replace all quantities with their space-partitioned counterparts to finally retrieve
\begin{equation} 
\mathcal{E}(u^*) := \int_{\Omega_3} \frac{D(\mathbf{f}^*,\mathbf{w}^*,u^*) }{\sum_i w_i^* + \gamma} + \lambda S(\nabla u^*)  \hspace{0.1cm} d \mathbf{x} ,
\label{eq:energy}
\end{equation}   
and instead solve for $u^*$ with a small $\gamma$ in the normalizer to avoid division problems for unseen voxels. To optimize Equation \ref{eq:energy}, we determine the steady state of our PDE
\begin{equation} 
\frac{\partial \mathcal{E}}{u^*} = \lambda \divergence( S_{\nabla u^*}(\nabla u^*) ) - \frac{D_{u^*}(\mathbf{f}^*,\mathbf{w}^*,u^*) }{\sum_i w_i^* + \gamma}.
\end{equation} 

Note that we, strictly speaking, optimize a new $u^*$ in each iteration since we constantly change the structure of our iterate. Nonetheless, this showed to be not a problem in practice since we observed a proper convergence in every case. We will now focus on clarifying how we evaluate above terms and how to conduct the actual optimization in the Octree.


\begin{figure}
\begin{floatrow}

\ffigbox{\includegraphics[width=6cm]{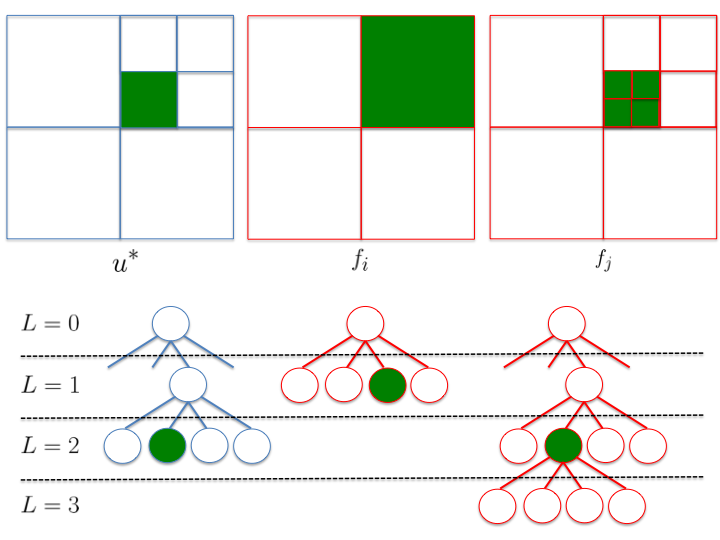}}
{	\label{fig:data_term}	\caption{Data term computation. Standing at the green node $n$ in $u^*$ at level 2, we query all TSDFs at the same spatial location. Either the level is not available ($f_i$) in which case we fetch the node that spatially subsumes $n$ or the level is the same/deeper ($f_j$) and we fetch the pre-computed value at the same level.}}	
			
\ffigbox{ \includegraphics[width=6cm]{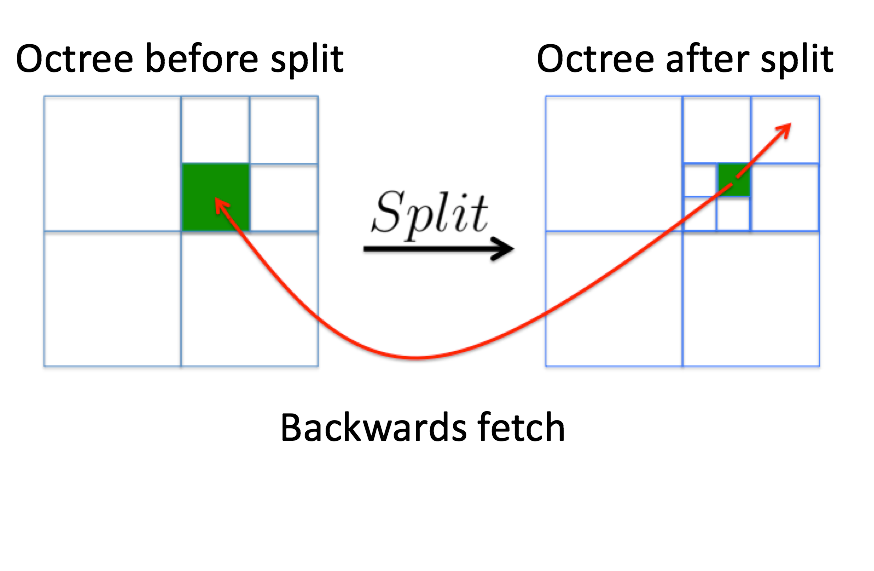}}
{	\label{fig:sweep}	\caption{Regularizer computation. The red arrows symbolize which nodes are needed to get the divergence for the green node on the right side. Fetching forward differences is always possible during the recursive run. For backward differences we avoid numerical errors by storing the node value before its split.}}

\end{floatrow}
\end{figure}

\paragraph{Optimization on the Octree} 

We conduct the optimization by having at all times only one version of $u^*$ in memory and adjusting the structure while we recursively traverse into each node of $u^*$. This means that instead of integrating point-wise over the volume, we start from the root and run along the tree while conducting our computations/restructuring on it before proceeding to the next node in the volume in the same pass. To mathematically facilitate the notation, we allow our TSDFs to take nodes as arguments.

For the repartitioning during optimization, we retrieve the gradient descent update $\Delta u^*_t$ at iteration $t$ for $u^*_t$. Since the update might encompass larger numerical changes, we need to reflect this by restructuring $u^*_{t+1}$ before applying the update such that no crucial information is lost. In contrast to \cite{Losasso2004,Losasso2006} we do not refine the cube resolution of the Octree simply based on their spatial distance to the narrow band but rather refine them based on numerical values.
For an Octree node $n$ during our run,  we compute, together with a gradient step size $\xi$,  the new value $\Delta n = u^*_t(n) + \xi \cdot \Delta u^*_t(n)$ and decide for the repartitioning together with a splitting threshold $\tau_s$, a joining threshold $\tau_j$ and two rules: 

\begin{itemize}

\item if $n$ is a leaf of the Octree and  $| \Delta n | < \tau_s$, we split and recurse into the children

\item else $n$ is not a leaf of the Octree and we check if  $| \Delta n | > \tau_j$. If this holds, we recursively conduct the same check for the children and if successful, we join these children only if furthermore they all hold values of equal sign. Otherwise, an implicit surface passes through these nodes and joining them might rupture it.
\end{itemize}

In order to compute node-wise expressions in our Octree-TSDFs, we simply fetch the corresponding values from the tree nodes that represent the subvolume at this position: if $u^*(n)$ resides at tree level $L$, we either fetch the pre-computed corresponding values $f_i(n)$ and $w_i(n)$ at level $L$, if the level exists, or take the closest leaf $n'$ that spatially subsumes $n$ (see Figure 4). The $f_i$ can be efficiently queried while moving alongside $n$ in each Octree.


For the more complex quantity, the gradient $\nabla u^*$, we use forward differences which need to be fetched in a neighborhood around each node $n$. This can be easily accomplished during the same pass since we can spatially look-up all nodes ahead of $n$ which have not been touched yet. To compute the divergence, we also need to be able to compute backward differences. In our approach we want to be fast and therefore want to accomplish one optimizer iteration in one pass through the Octree. Thus, we immediately restructure all visited nodes and would therefore induce numerical errors if we fetch backwards during the same pass. To remedy that we store for a node its value before splitting such that the computation is proper (see Figure 5). Note that this is not applicable when joining a node since it would need to carry a history of all its children values. However, due to our splitting rules, joins never happen at interfaces and we thus can discard this otherwise problematic issue.

\subsection{Meshing and coloring}
As a final step, we apply Marching Cubes to extract the 0-level isosurface to retrieve a mesh. To compute the coloring of the resulting mesh, we weigh for each vertex the reprojected colors according to the dot product between its normal and every camera view vector, similar to \cite{Whelan2013}. Furthermore, to supply color to unseen parts, we iteratively propagate the colors along triangle boundaries and blend the final color in respect to the neighboring, colored vertices and their dot products.

\section{Evaluation}
The method was implemented in C++ and the experiments were conducted on a CPU with 32GB of RAM. We ran our experiments on different kinds of data: firstly, we synthetically rendered 31 views of a sphere to measure our loss in accuracy on perfect, noise-free data. Secondly, we acquired sequences of four objects (each consisting of 26 frames) with a commodity RGB-D sensor (Carmine 1.09) to assess the geometrical quality we can achieve with low-cost devices. Lastly, we acquired 24 frames of a turbine blade taken with an industrial high-precision depth sensor (GIS) that provides micrometer precision. Since we have a CAD model of the turbine we evaluate how accurate we can reconstruct real-life objects with state-of-the-art depth sensing technology which is important for manufacturing verification. For comparisons with dense results, we implemented the method from \cite{Kehl2014}.

For the experiments we found that running the optimization for $100$ iterations with an initial gradient step size $\xi=0.1$ and halving it every $20$ iterations was sufficient to converge to good solutions for any object. Furthermore, we fixed $\eta= 2cm$ but set the metric voxel size $s_v$ and the uncertainty factor $\delta$ depending on the data source. For the  synthetic data, we set $s_v := 1mm$ and $ \delta := 0.1mm$, for the Carmine dataset $s_v :=1.5mm, \delta := 2mm$ and for the GIS data $s_v = 0.5mm, \delta := 0.8mm$.

\begin{figure}
	\centering
	\includegraphics[width=3cm,height=2.5cm]{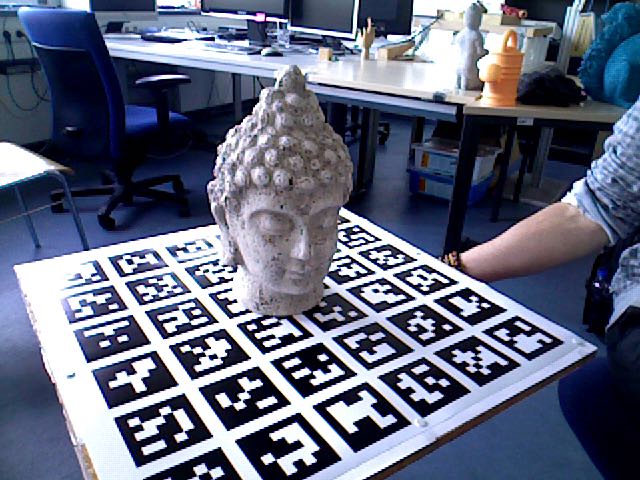}
	\includegraphics[width=3cm,height=2.5cm]{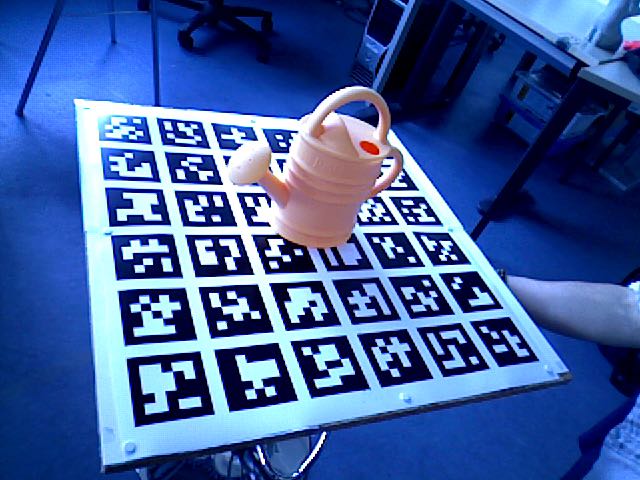}
	\includegraphics[width=3cm,height=2.5cm]{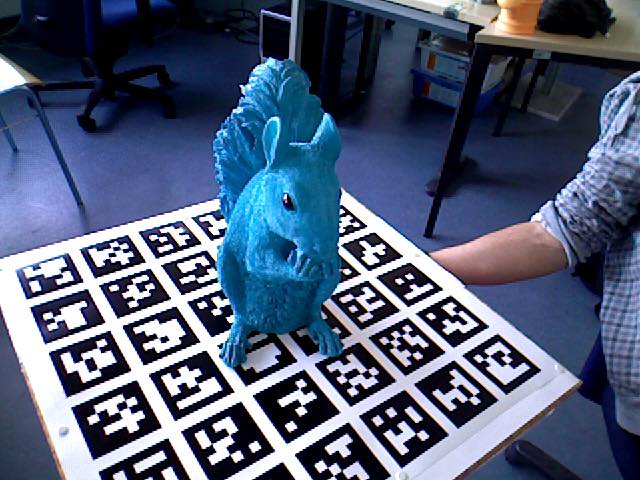}
	\includegraphics[width=3cm,height=2.5cm]{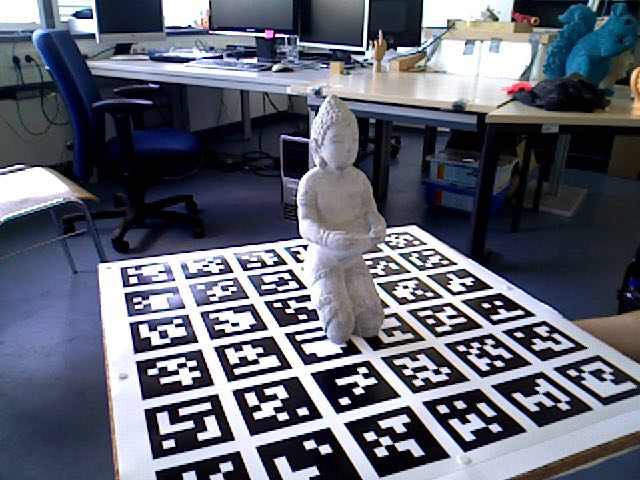} 
	 \\
	 \vspace{0.1cm}	
	 \includegraphics[width=3cm,height=2.5cm]{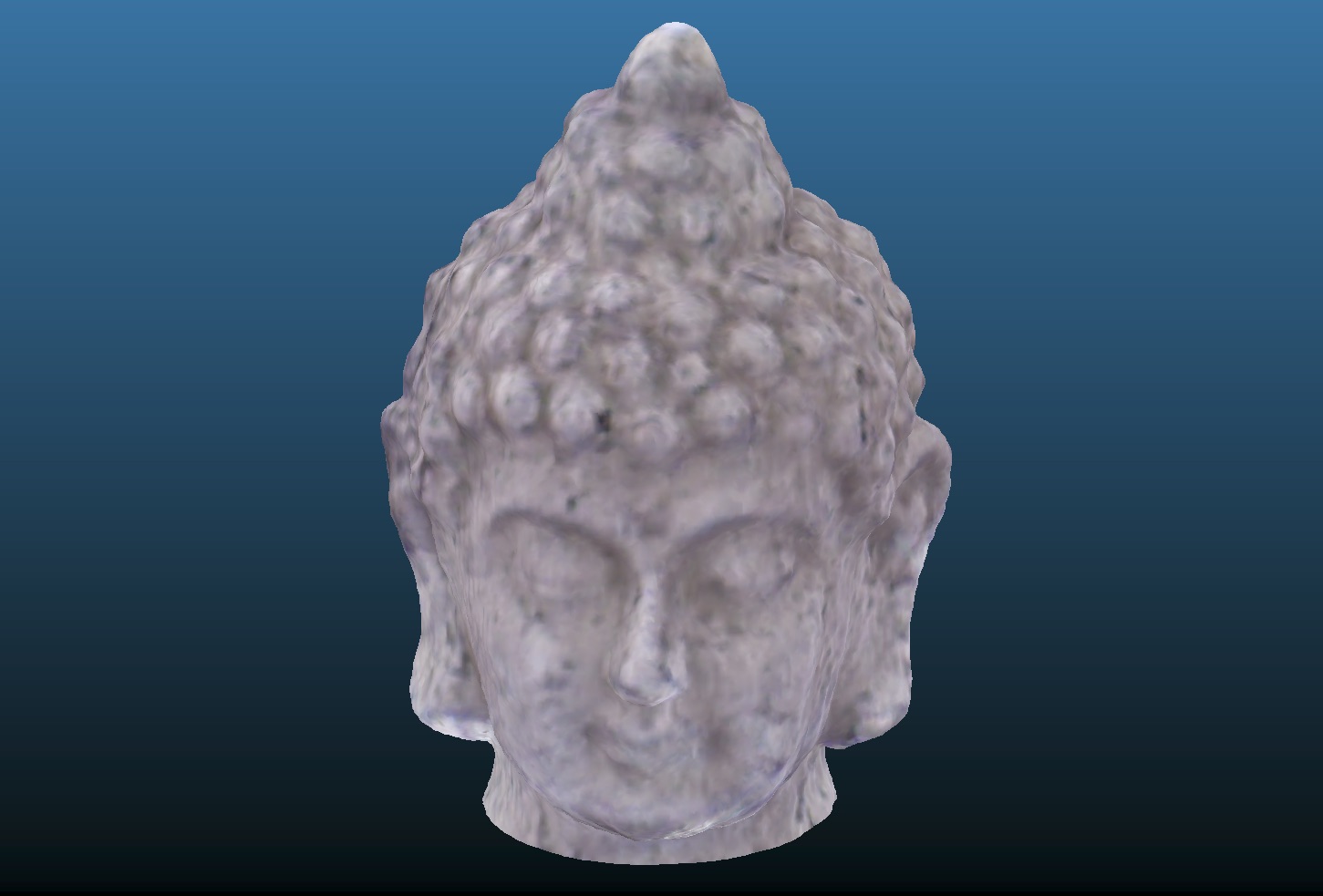}
	 	\includegraphics[width=3cm,height=2.5cm]{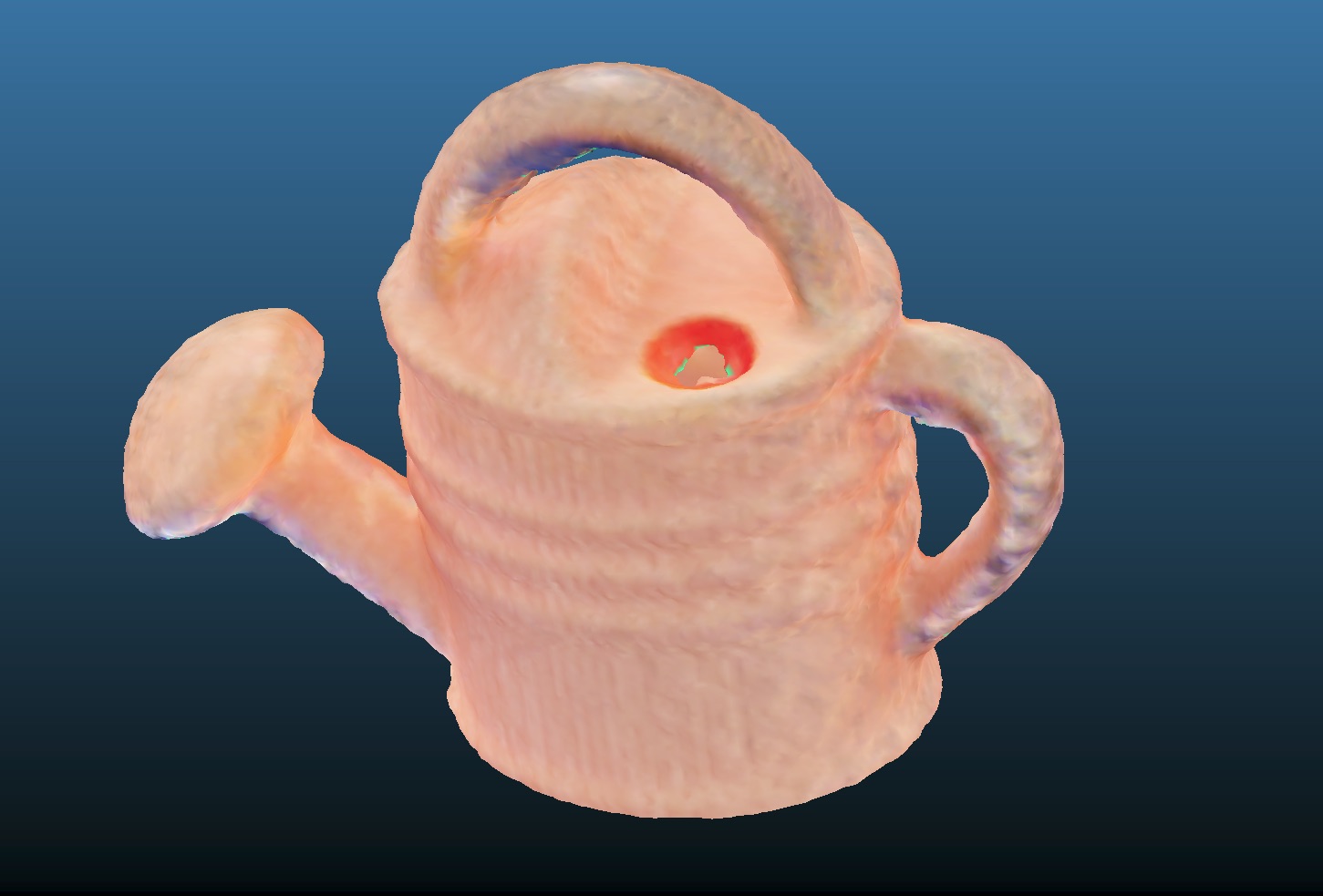}
	 	\includegraphics[width=3cm,height=2.5cm]{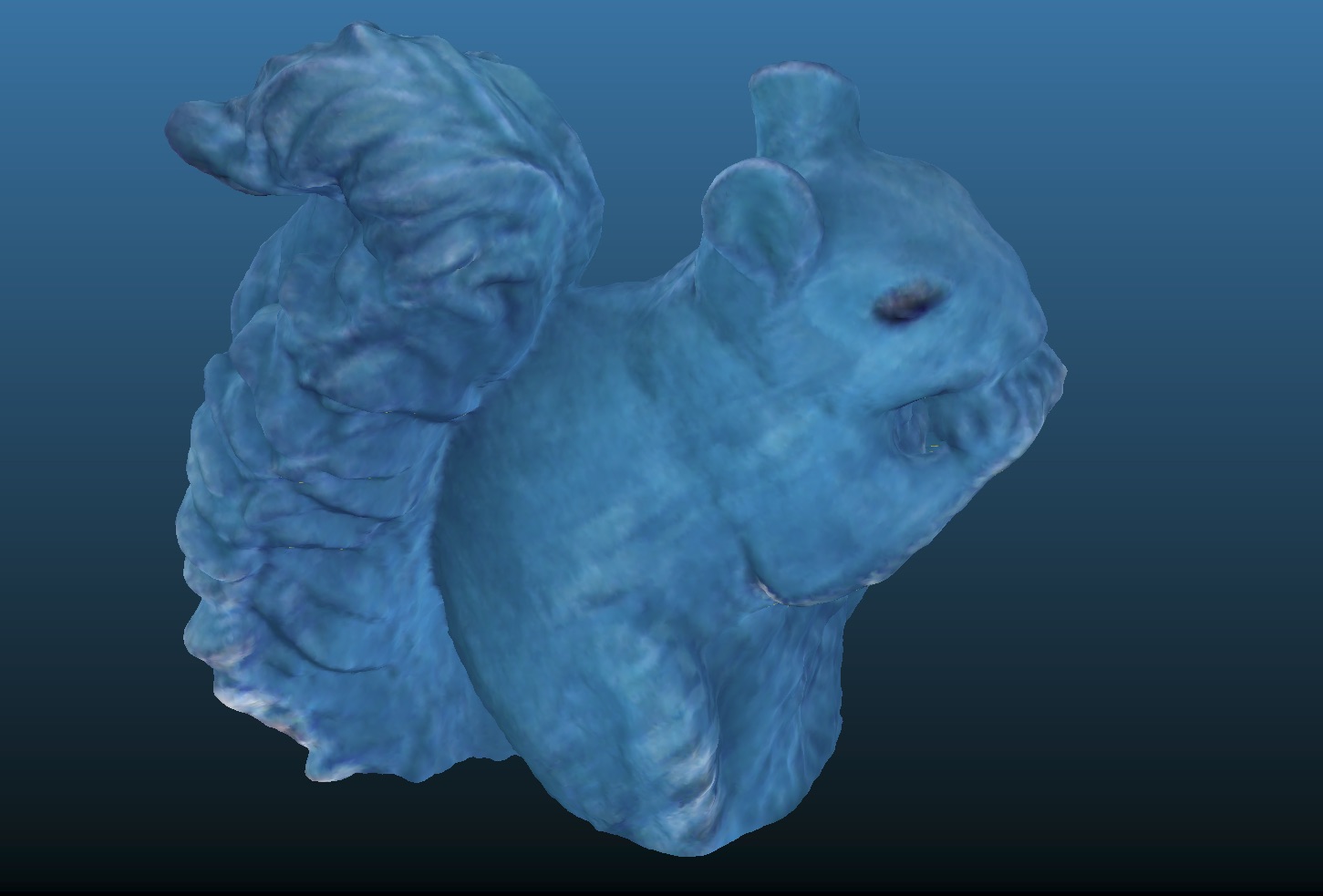}
	 	\includegraphics[width=3cm,height=2.5cm]{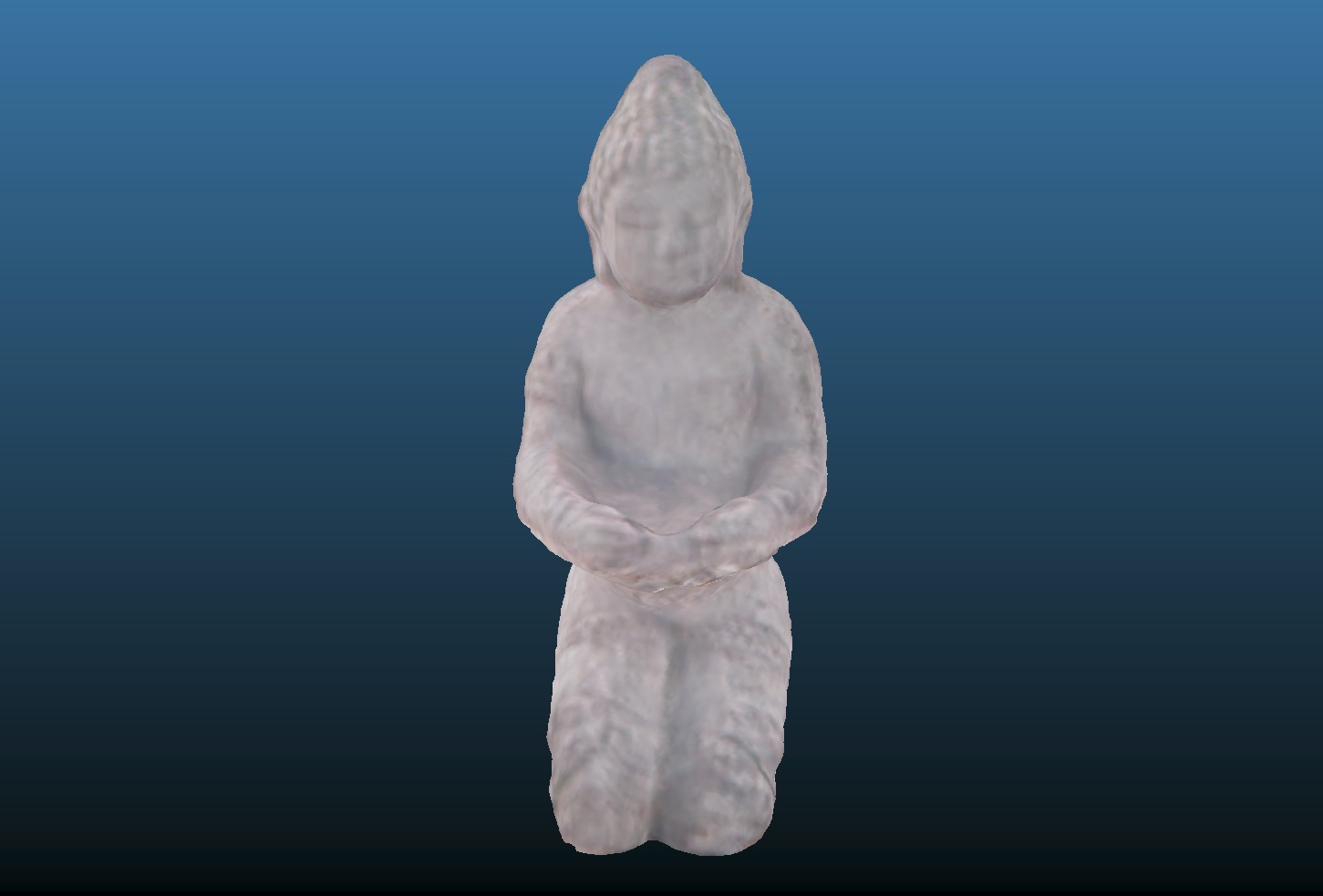} 
	 	\\
	 	\vspace{0.1cm}	
		 \includegraphics[width=3cm,height=2.5cm]{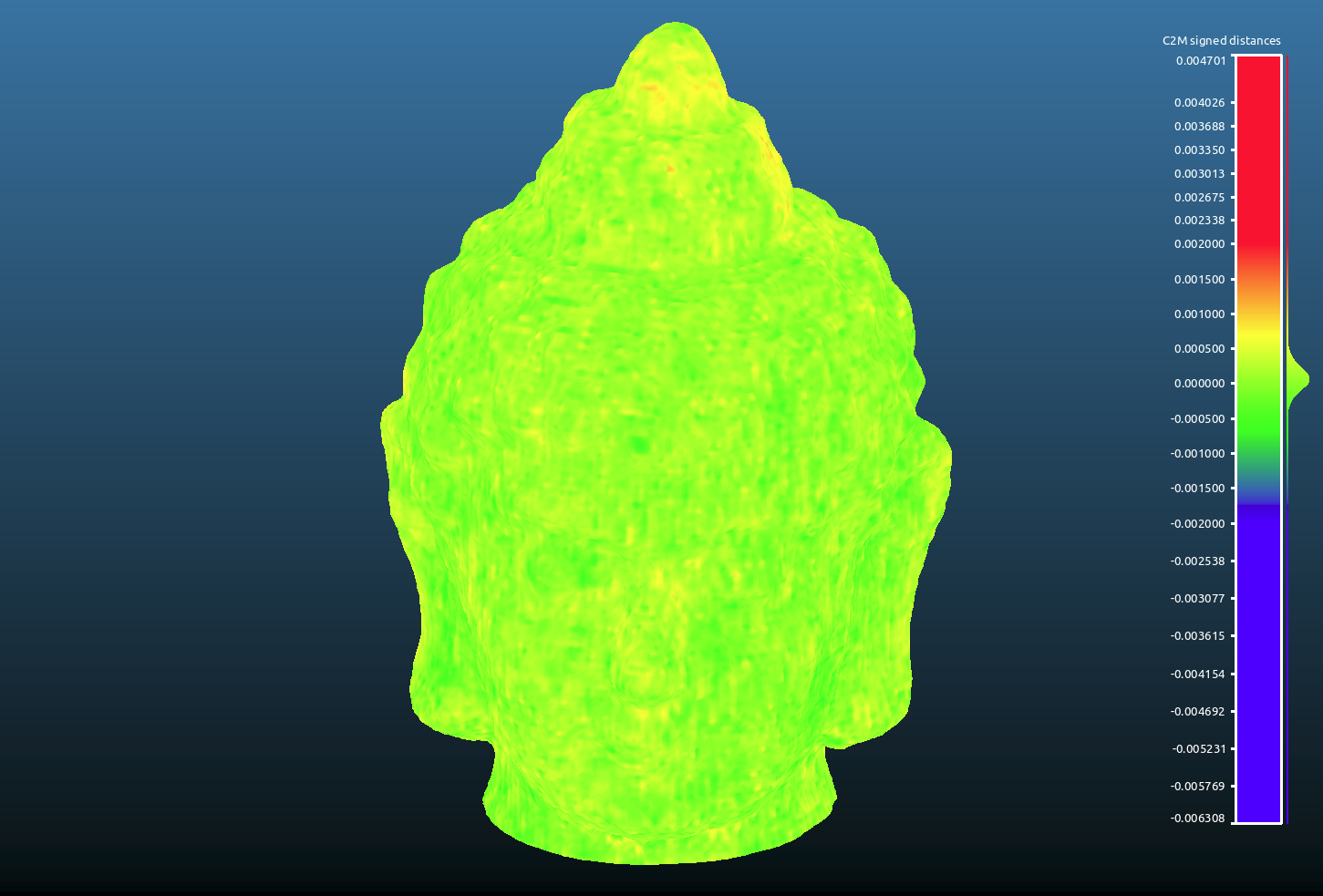}
	 	\includegraphics[width=3cm,height=2.5cm]{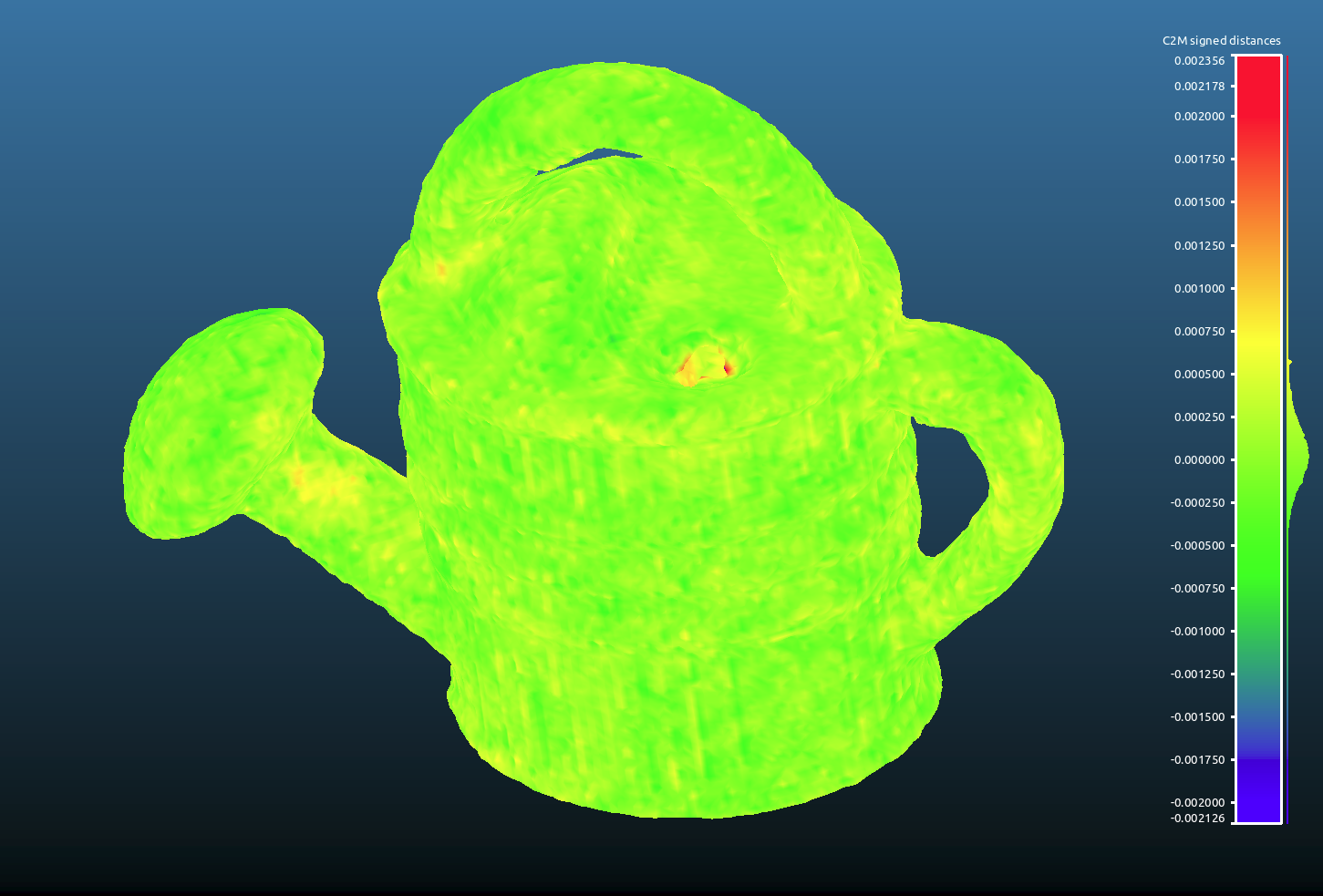}
	 \includegraphics[width=3cm,height=2.5cm]{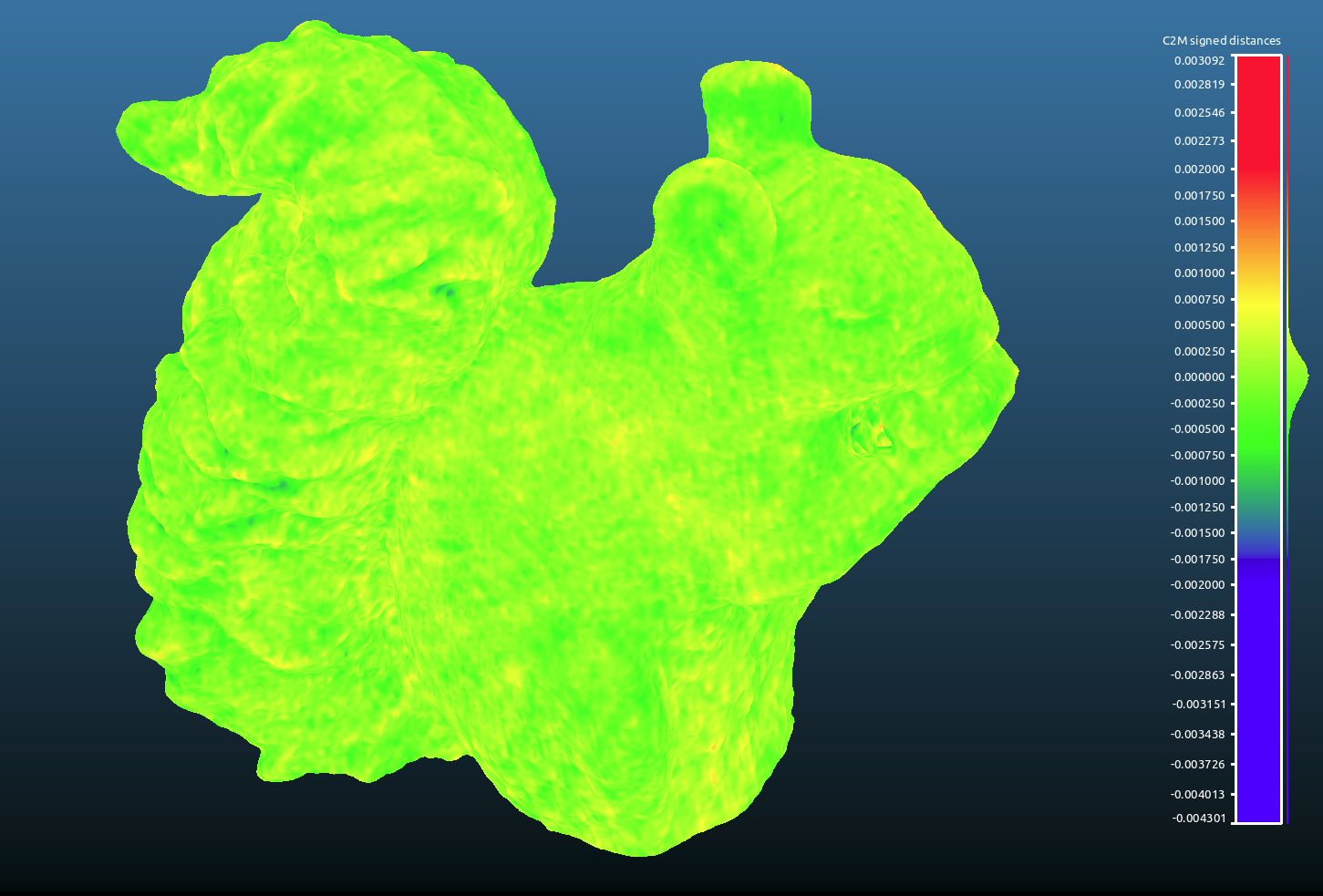}
	 \includegraphics[width=3cm,height=2.5cm]{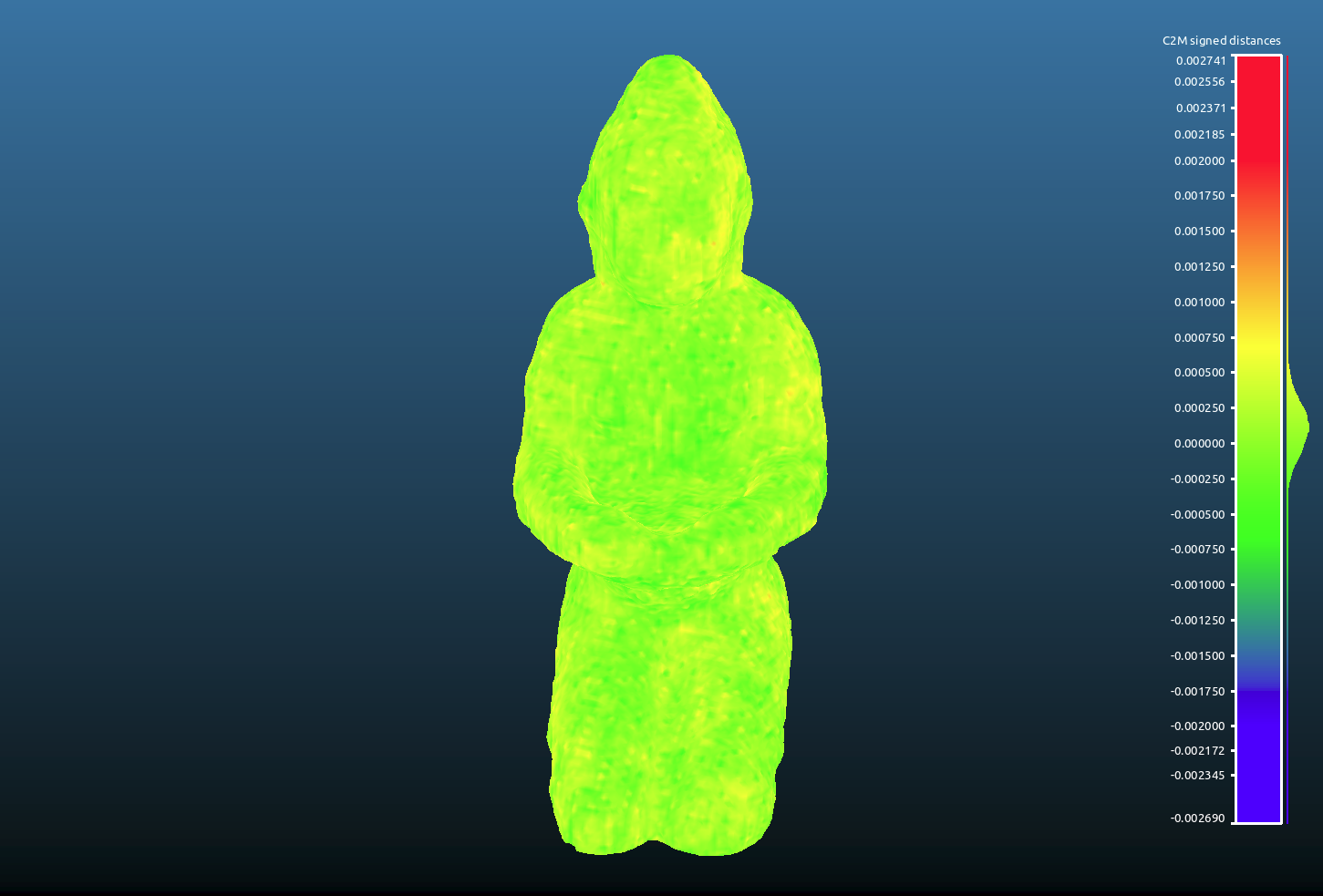} 
	 		 	
	\caption{The four objects acquired with the Carmine 1.09 sensor (top) together with their dense reconstructions (center) and their vertex-wise difference to their Octree-reconstructed pendants (bottom). The minimum voxel size was set to $1.5$mm and $\lambda=0.3$ for all objects and both methods. The saturation of the difference coloring has been set to $\pm 2mm$. Obviously, the error induced by our approach stays bounded within the specified voxel size of $1mm$.}
	\label{fig:carmine}
\end{figure}

\begin{figure}
	\centering
	\includegraphics[width=4cm]{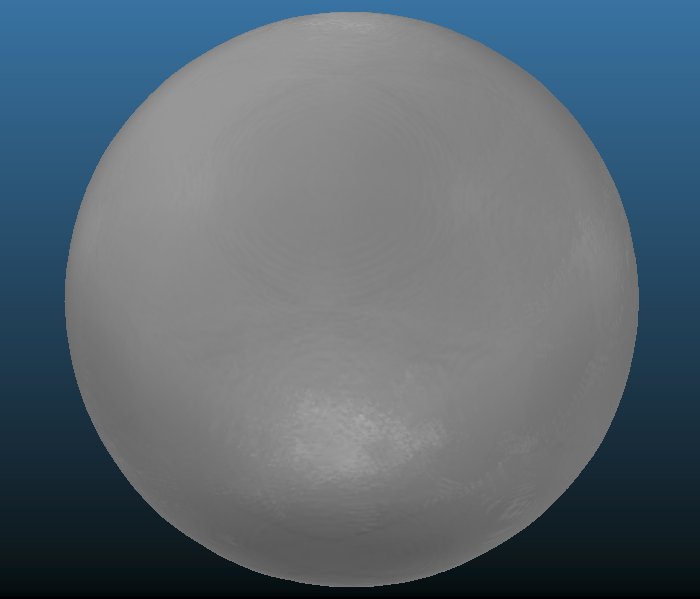}
	\includegraphics[width=4cm]{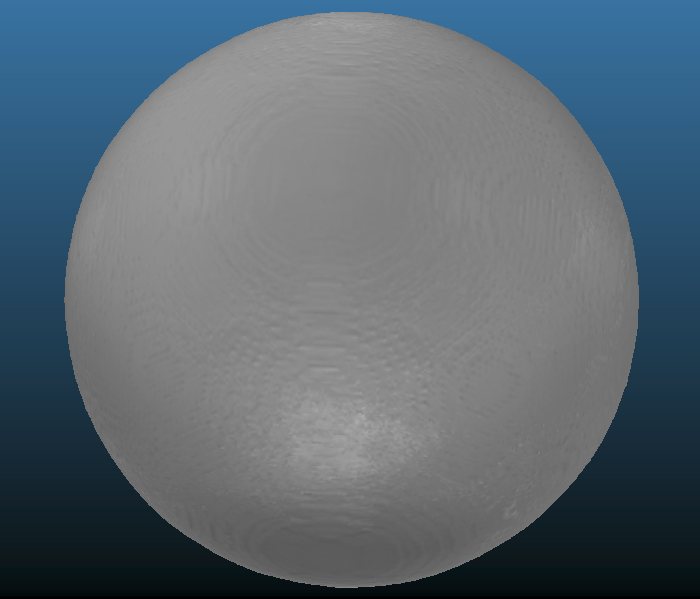}
	\includegraphics[width=4cm]{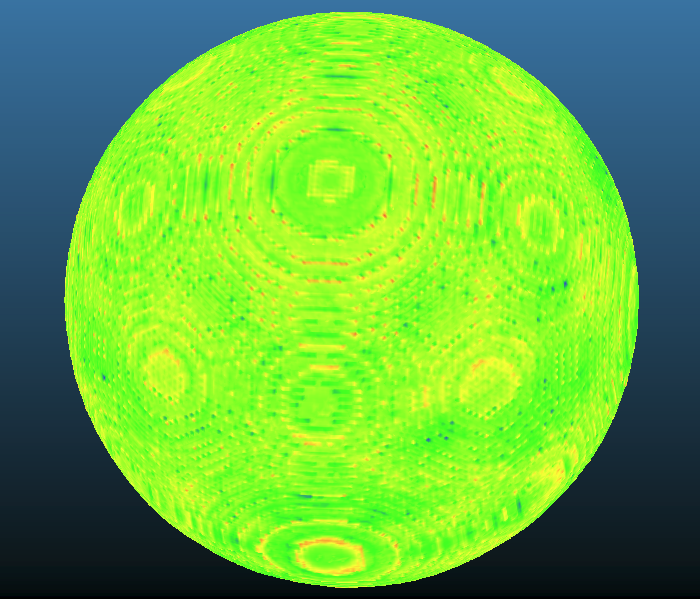}
	\caption{Left: Dense result. Middle: Octree result. Right: Vertex-wise difference.}
	\label{fig:sphere}
\end{figure}

\begin{figure}
	\centering
		\includegraphics[width=4cm,height=4cm]{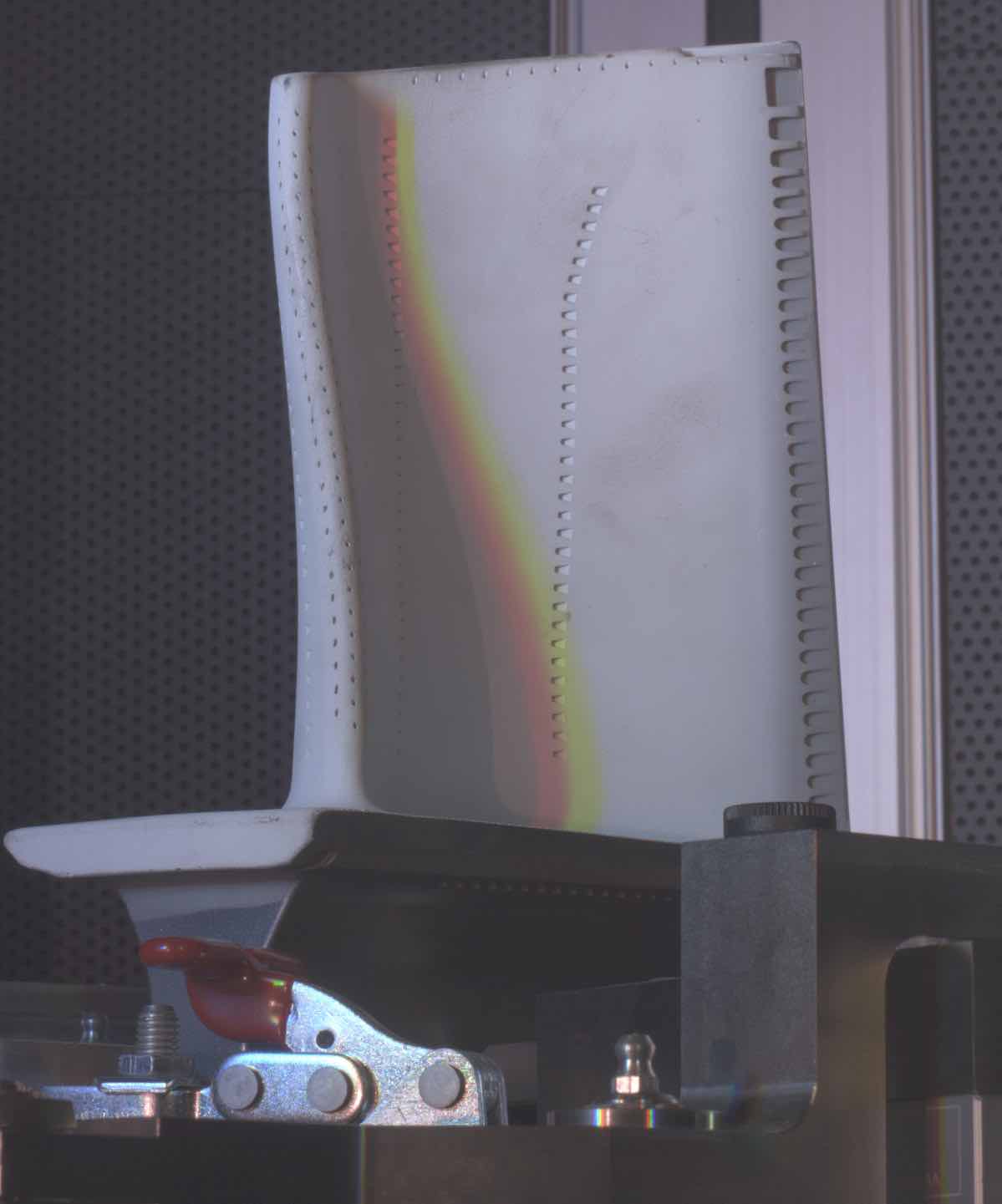}
		\includegraphics[width=4cm]{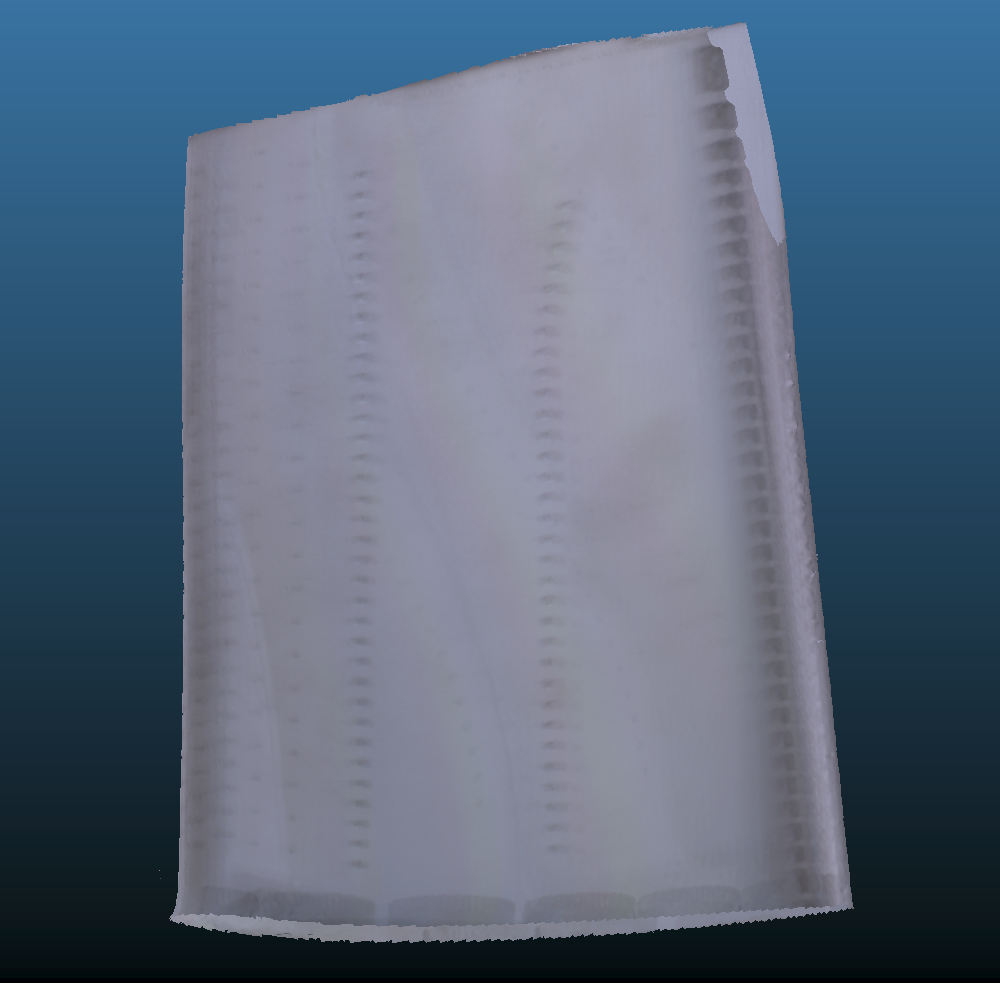}
		\includegraphics[width=4cm]{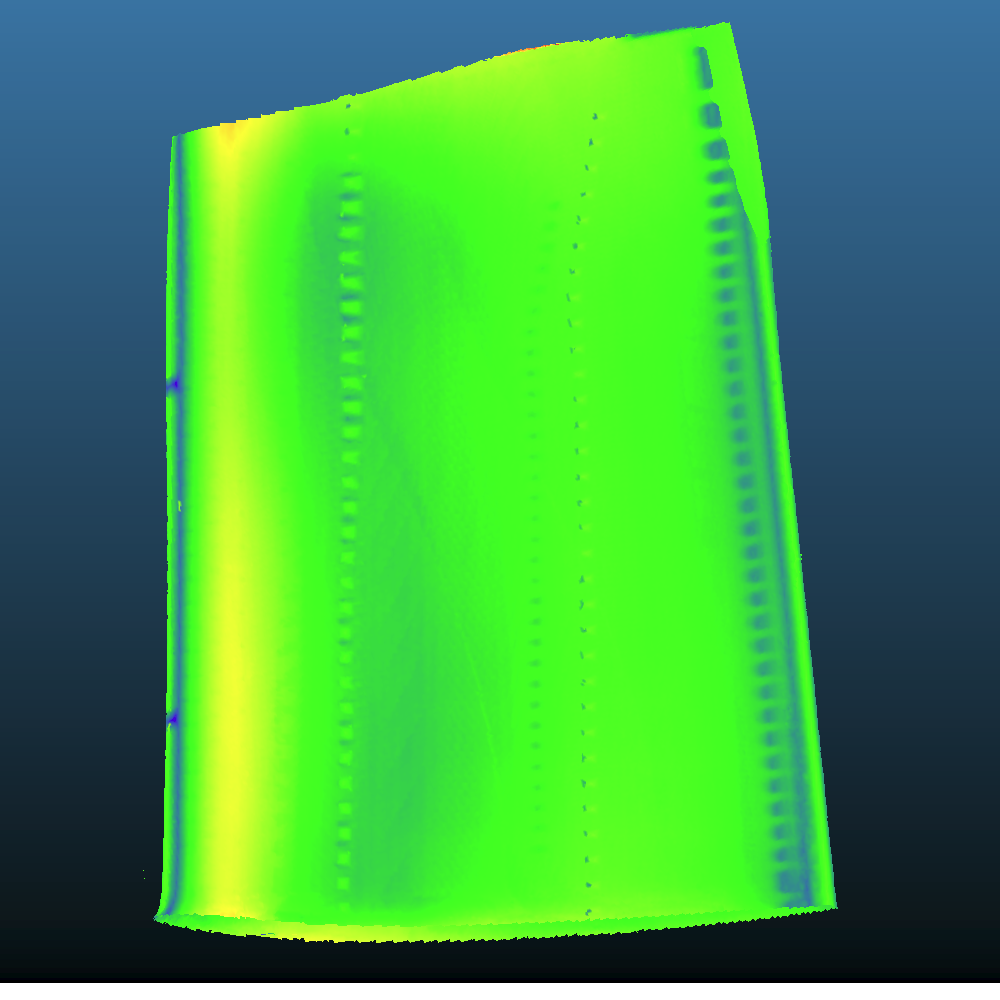}
		\caption{'Turbine' reconstruction. From left ro right: One frame from the sequence, the Octree-reconstructed 3D mesh and the difference to the CAD model with mean error of $0.22mm$ and standard deviation of $0.5mm$. Most errors accumulate at the sharp edge on the left as well as the small indents on the right which where smoothed during optimization.}
	\label{fig:turbine}
\end{figure}

For the Carmine sequences (Figure \ref{fig:carmine}), we constantly retrieve very accurate solutions that are on par with their dense versions. For a more quantitative comparison, we also compare our reconstructions of the 'sphere' and the 'turbine' to their groundtruth models (Figures \ref{fig:sphere} and \ref{fig:turbine}). To compute the vertex-wise difference, we find for each vertex of one reconstruction the closest point of the other reconstruction and compute their distance. The synthetic sphere is nicely reconstructed and the mean error of $0.012mm \pm 0.070mm$ is virtually negligible. Also the 'turbine' reconstruction was very accurate with a mean error of $0.22mm \pm 0.50mm$ which is inside tolerable limits.

\subsection{Memory consumption and runtime}
 
\begin{figure}
	\centering
	\includegraphics[width=4cm]{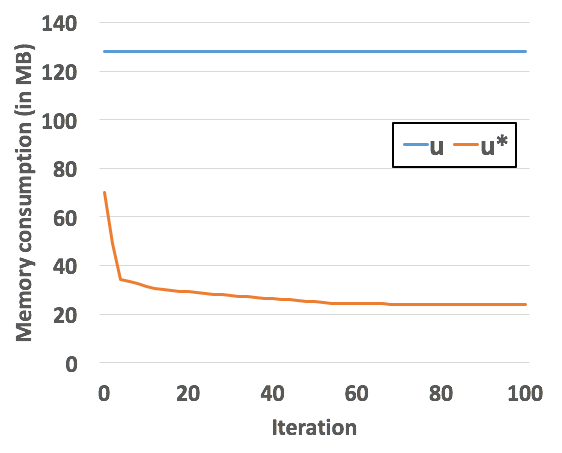}
	\hspace{0.25cm}
	\includegraphics[width=3.3cm]{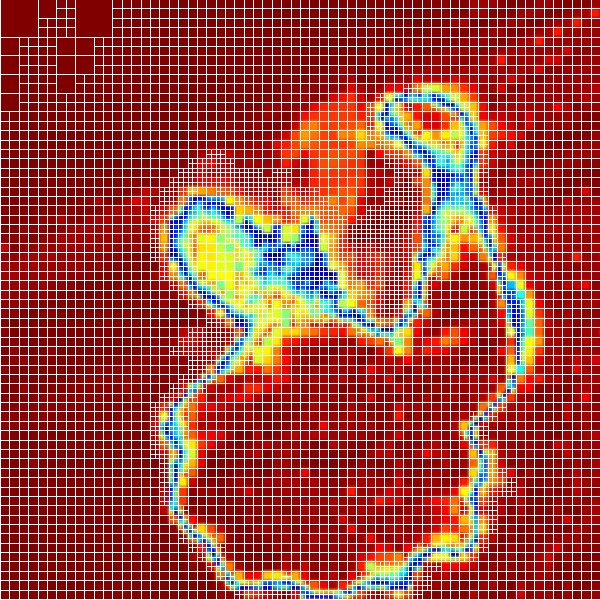}
	\includegraphics[width=3.3cm]{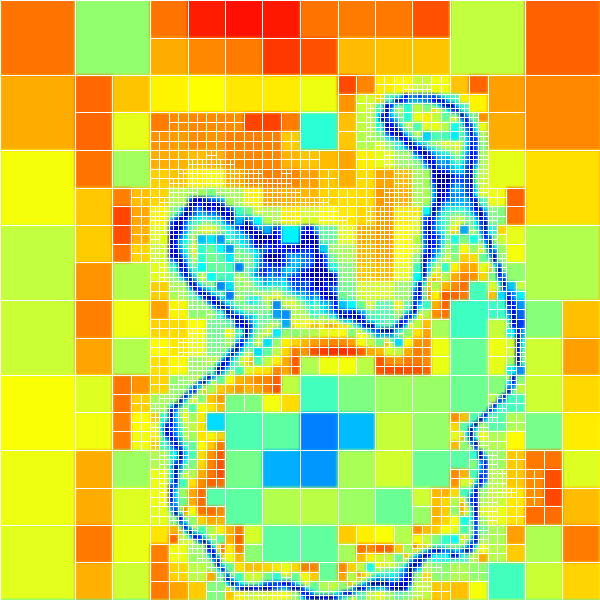}
	\caption{Left: Memory usage of the iterate $u$ and $u^*$ during the optimization for the 'head' sequence. The usage goes down quickly for the Octree-variant as the surface evolves in the TSDF, leading to many block joins. Center/Right: Slicing through $u^*$ at iterations 1 and 100. }
	\label{fig:memory_u}
\end{figure}

Each dense TSDF stores for each voxel the actual distance value and the weight as floats. In comparison, the Octree-TSDF stores per block pointers to 8 children, its parent node and in addition to the above two floats another float value that represents the distance value just before splitting to allow for the computation of the divergence in the same pass. 
The total amount of memory needed to hold the data term is given in Table 1 and is drastically reduced with the Octree approach for all sequences. To give another interesting insight we plot the memory consumption of $u^*$ during the optimization in Figure \ref{fig:memory_u}. While in the dense approach each iterate $u_t$ is constant in memory, its Octree-variant quickly decreases its footprint after more and more blocks get joined. Analogously for the runtime in Table 2, our fast traversing technique allows us to also outperform the dense variants except for the 'squirrel'. While the runtimes for the dense TSDFs are dependent on the volume resolution, the geometry complexity is the main driver of the runtime for the Octree method since frequent repartitionings inbetween iterations can lead to a runtime penalty.

\subsection{Split and join}

As stated, we apply split and join rules according to thresholds $\tau_s$ and $\tau_j$. Since these values govern the structural density of our Octrees, a careful choice is important to uphold the geometrical accuracy. Splitting early creates a finer partitioning and can lead to unnecessarily high runtime and memory demands while joining too early can result in larger numerical errors since smaller gradient increments get discarded. To have a visual feeling of the impact, we show the fusion of some Carmine frames taken of a 3D print of the Stanford bunny in Figure 11. Apart from the first case that virtually hinders nodes from splitting and leading to massive Octree-artifacts with $\tau_s = 0$, the geometry suffers less from quantization with an increasing $\tau_s$ since it steers how fine the Octree describes the area around the narrow band. Conversely, with a higher $\tau_j$ we can delay the joining of nodes and thus have the same effect on the area around the band, but from the opposite side, and with a value of $\tau_j>1$ actually disabling join operations.

\begin{figure}
\begin{floatrow}
\capbtabbox{
 \begin{tabular}{@{}c|c|c@{}}
       & SDF Size/Res.  & Octree Size/Res.  \\
      \hline
      sphere &  $3,968$ MB / $256^3$ & $257$ MB / $256^3$ \\
      \hline 
      statue	&  $3,072$ MB / $256^3$ & $683$ MB / $256^3$ \\
      head 		& $3,072$ MB / $256^3$	& $806$ MB / $256^3$ \\
      squirrel &  $3,072$ MB / $256^3$	&  $689$ MB / $256^3$ \\ 
      can 		&  $3,072$ MB / $256^3$	&  $565$ MB / $256^3$ \\ 
      \hline
      turbine & $24,576$ MB / $512^3$	&  $2,192$ MB / $512^3$
    \end{tabular}
\label{table:memory} 
}{\caption{Memory consumption/volume resolution for the input TSDFs $f_i$ for each sequence.}}

\capbtabbox{
 \begin{tabular}{@{}c|c|c@{}}
       & Dense  & Octree  \\
      \hline
      sphere &  3.9 & 2.8 \\
      \hline 
      statue	&  3.7 & 2.6 \\
      head 		& 3.7 & 3.4 \\
      squirrel &  3.7 & 4.7 \\
      can 		&  3.7 & 3.2 \\
      \hline
      turbine & 26.1 & 9.5 \\
    \end{tabular}
\label{table:runtime} 
}{\caption{Runtime (in minutes) for the optimization.}}			

\end{floatrow}
\end{figure}

\begin{figure}
\begin{floatrow}

\ffigbox{	
\includegraphics[width=5cm]{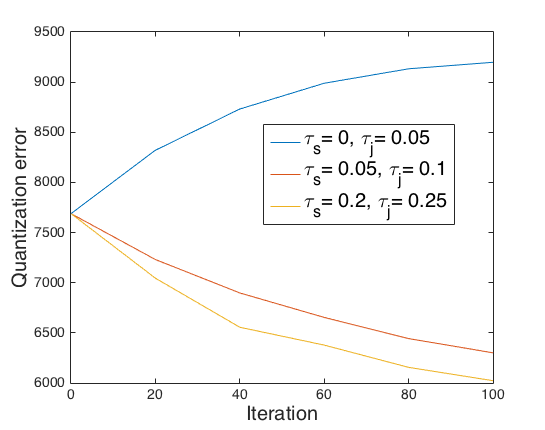}
}
{	\label{fig:error}	\caption{The quantization error with three configurations. The higher the thresholds, the smaller the approximatation error.}}	
			
\ffigbox{ 
	\includegraphics[width=2cm]{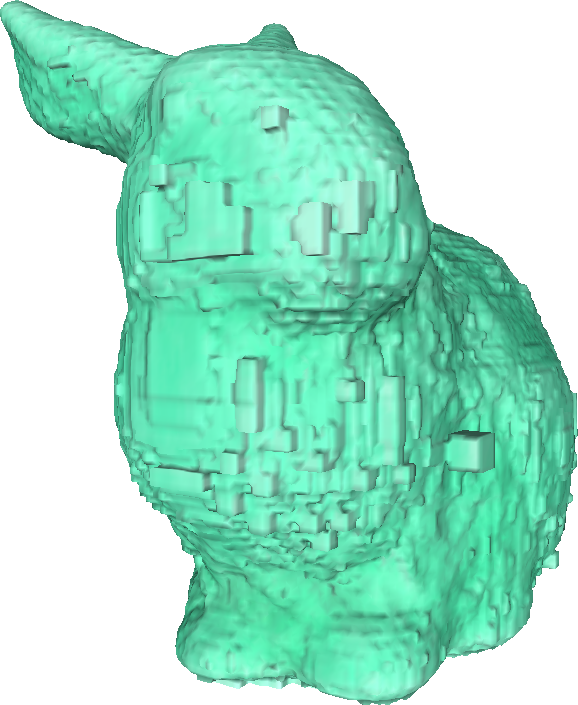} 
	\includegraphics[width=2cm]{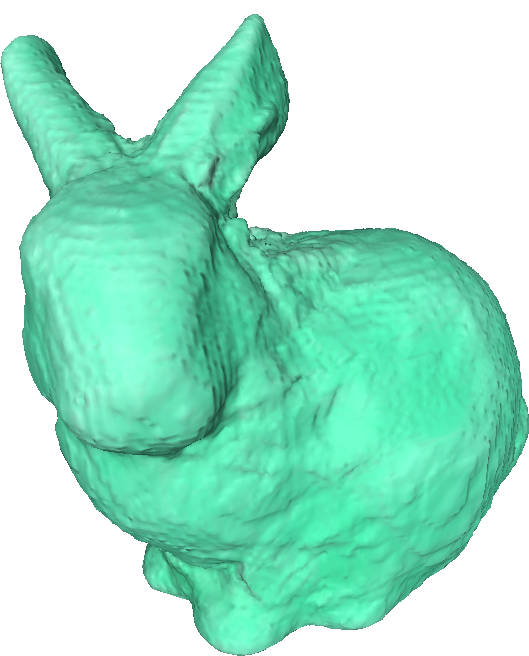} 
	\includegraphics[width=2cm]{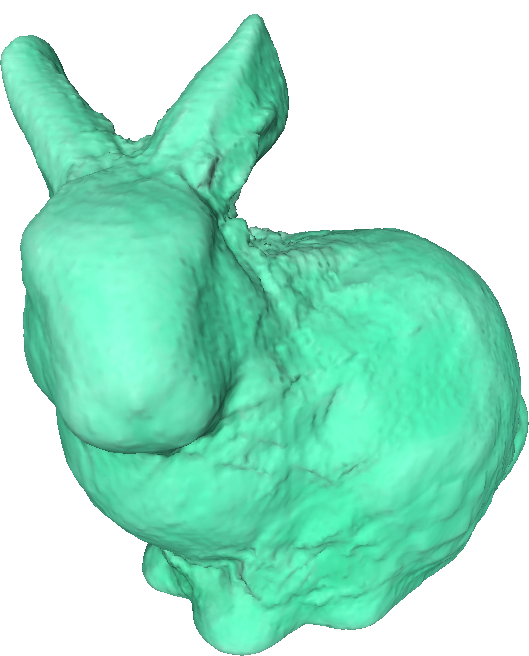} 
}
{	\label{fig:bunny}	\caption{The quantization effect of the two join/split thresholds $\tau_s$ and $\tau_j$ for the three configurations from the left.}}

\end{floatrow}
\end{figure}

The quantization error between a dense $u$ and its induced Octree-version $u^*$ is 
\begin{equation} 
\sum _{n \in \leafs(u^*)} \bigg \vert u^*(n) - \frac{1}{\Omega_3(n)}\int_{\Omega_3(n)}  u(\mathbf{x}) d \mathbf{x} \bigg \vert
\label{eq:error}
\end{equation}  
and in the ideal case it should be zero for each iterate pair $(u_t,u^*_t)$ during optimization. As can be seen in Figure 10, the error decreases with higher values of both $\tau_s$ and $\tau_j$ whereas in the special case of $\tau_s=0$, the error grows larger due to the lack of splitting, leaving the surface heavily artifacted while the dense version gets smoother.

\section{Conclusion}

To our knowledge, we are the first to present an approach towards variational range data fusion by partitioning the solutions with a dynamic Octree structure. We have shown how to efficiently conduct restructuring based on iterative node-wise updates of the supplied PDE and that the achieved results are geometrically accurate on multiple datasets and nearly identical to their dense counterparts while being more efficient overall. It would be interesting to investigate further into different ways of computing the required quantities, alternative split/join decisions as well as other solving techniques (e.g. primal-dual \cite{Chambolle2011a}). 
\paragraph{Acknowledgements} The authors would like to thank Toyota Motor Corporation for supporting and funding this work.  

\bibliography{egbib}

\end{document}